# Mask wearing object detection algorithm based on improved YOLOv5★

Peng Wen, Junhu Zhang*, HaiTao Li

*College of Information Science and Technology, Qingdao University of Science and Technology ,Qingdao, Shandong, China*

ABSTRACT

Wearing a mask is one of the important measures to prevent infectious diseases. However, it is difficult to detect people's mask-wearing situation in public places with high traffic flow. To address the above problem, this paper proposes a mask-wearing face detection model based on YOLOv5l. Firstly, Multi-Head Attentional Self-Convolution not only improves the convergence speed of the model,but also enhances the accuracy of the model detection. Secondly, the introduction of Swin Transformer Block is able to extract more useful feature information, enhance the detection ability of small targets, and improve the overall accuracy of the model. Our designed I-CBAM module can improve target detection accuracy. In addition, using enhanced feature fusion enables the model to better adapt to object detection tasks of different scales. In the experimentation on the MASK dataset, the results show that the model proposed in this paper achieved a 1.1% improvement in mAP(0.5) and a 1.3% improvement in mAP(0.5:0.95) compared to the YOLOv5l model. Our proposed method significantly enhances the detection capability of mask-wearing.

## 1. Introduction

From the global spread of the COVID-19 pandemic to cross-infection by various mutated strains, wearing masks scientifically and consciously is one of the most effective ways to prevent the spread of the virus. However, there are still some people who lack scientific awareness, and the phenomena of not wearing masks or not wearing them properly in public places are still very serious[1]. Nowadays, the situation of wearing masks in public places mostly relies on manual inspection, which is not only costly and time-consuming, but may also lead to cross-infection among person. In order to promote global health, it is particularly important to develop a mask-wearing detection system for public places[2]. Artificial intelligence technology has shown excellent performance in various fields. The use of object detection methods in artificial intelligence technology can help relevant departments to quickly and accurately detect the mask-wearing situation of pedestrians.

Currently, researchers have conducted relevant studies on whether mobile individuals wear masks. Lin Sihai et al[3] improved the SSD algorithm by using ResNet50 instead of VGG-16 in the feature extraction part, and adding the SENet mechanism in the feature fusion part. The improved algorithm showed an increase in accuracy by 4.1%, 9.9%, and 5.3% respectively at IOU of 0.5, 0.75, and 0.50:0.95. However, the overall accuracy of the network is still relatively low. Wang Yifei et al.[4] used the YOLOv4 network and introduced the GhostNet convolutional neural network into the model, which not only reduced the model parameters and computation, but also improved the model accuracy by 5.52%, reaching 93.21%. Akhil Kumar et al.[5] improved the micro YOLOv4 network by introducing the SPP model after the backbone to enrich feature extraction. At the same time, in order to detect small targets, a YOLO detection layer was added. The detection accuracy was improved by 14.05% compared to the original model in terms of the average precision (AP). Pham ThiNgot et al. [6]introduced the CA attention mechanism into the YOLOv5 model, and improved the network's mAP(0.5) to 96.8%, surpassing the detection accuracy of YOLOv6, YOLOv7, and YOLOX[7] models. In the case of mask wearing detection, YOLOv5 showed good performance in terms of accuracy, so this paper chose to optimize this model.

*Junhu zhang: Tel.: 13658675791;
*e-mail:* wen_peng.de@qq.com (Peng Wen)
jzhang@qust.edu.cn(Junhu Zhang)

## 2. Related Work

Object detection is one of the mainstream tasks in computer vision, which mainly involves predicting the classification and location of objects. Currently, commonly used object detection algorithms can be roughly divided into three categories: traditional object detection algorithms, two-stage object detection algorithms, and one-stage object detection algorithms[8].

Traditional object detection algorithms can be divided into three steps: selecting candidate regions, extracting features from the candidate regions, and classification. Common feature extraction algorithms include HOG[9], SIFT[10] algorithms, etc. Finally, a classifier is used to classify the extracted features. The main classifiers used are SVM[11], AdaBoost[12] , etc. However, traditional object detection algorithms have the following two limitations: (1) poor recognition performance and low accuracy, (2) many redundant windows, high computational complexity, and high time cost.

Two-stage object detection algorithms have two steps: generating candidate regions and then classifying using deep learning algorithms. The main algorithms include R-CNN[13], Fast R-CNN[14], Faster R-CNN[15], R-FCN[16], Mask R-CNN[17], etc. Although these algorithms have high detection accuracy, they suffer from slow detection speeds and high computational complexity, which makes it difficult to achieve real-time object detection tasks.

One-stage object detection algorithms are regression-based detection algorithms, They don't need to generate candidate boxes in advance. Instead, they directly extract features, classify and locate objects using a convolutional neural network. The main algorithms include SSD[18], RetinaNet[19], and the YOLO series (YOLOv1[20], YOLO9000[21], YOLOv3[22], YOLOv4[23], YOLOv5[24], YOLOv6[25], YOLOv7[26], YOLOv8[27]), etc. Because they directly perform the classification and location tasks on the objects, one-stage object detection algorithms are faster than two-stage algorithms in terms of detection speed. Currently, one-stage object detection algorithms have become the most popular research focus in the field of object detection, and researchers are working to improve the accuracy of these algorithms while maintaining their speed.

In computer vision tasks, the use of attention mechanisms can help capture important regions of images more quickly and efficiently while ignoring unimportant features. Hu et al.[28] proposed SENet, which includes Squeeze and Excitation operations to learn channel relationships and reduce prediction errors by adding them to ResNet residual structures. Wang et al.[29] proposed the ECA module, which increases direct connections between channels and weights and uses a single FC layer of cross-channel interaction instead of dimension reduction, considering that only the channel mechanism leads to information loss, Sanghyun et al.[30] proposed the CBAM hybrid attention mechanism, which includes both channel attention (CAM) and spatial attention (SAM). The attention used concatenation to separately fuse the attention weights of channels and spaces and then concatenated to generate the final feature map. Extensive research has shown that using attention mechanisms correctly can improve model detection capabilities.

To detect the wearing of masks by people in public places, this paper proposes an improved algorithm based on YOLOv5l model. The algorithm replaces the first layer of the network with a Multi-Head Attention Self-Convolution module, which accelerates model convergence and enhances detection capabilities. Instead of the first C3 module, the Swin Transformer Block[31] is used to enhance the detection capabilities for small or dense objects, thereby improving the model's detection accuracy. Before feature fusion, the I-CBAM (Improved Convolutional Block Attention Module) attention mechanism, which is an improved version based on CBAM to improve overall accuracy. Finally, enhanced feature fusion is used to enable richer semantic and information exchange among the same levels in the network.

## 3. Model Introduction

### *3.1. YOLOv5*

YOLOv5(Ultralytics,2020) consists of 5 versions: YOLOv5n, YOLOv5s, YOLOv5m, YOLOv5l, and YOLOv5x. Although the "s" and "m" versions of YOLOv5 are known for their fast processing speed, they tend to have lower accuracy. The YOLOv5x version offers relatively higher detection accuracy, but it comes with increased network depth and width, which results in slower processing speed. Therefore, this paper chooses the relatively balanced YOLOv5l network for experimentation. The YOLOv5l network structure comprises three components: the feature extraction layer (Backbone), the feature fusion layer (Neck), and the prediction feature layer (Head). **Figure. 1** illustrates the network architecture of YOLOv5l.

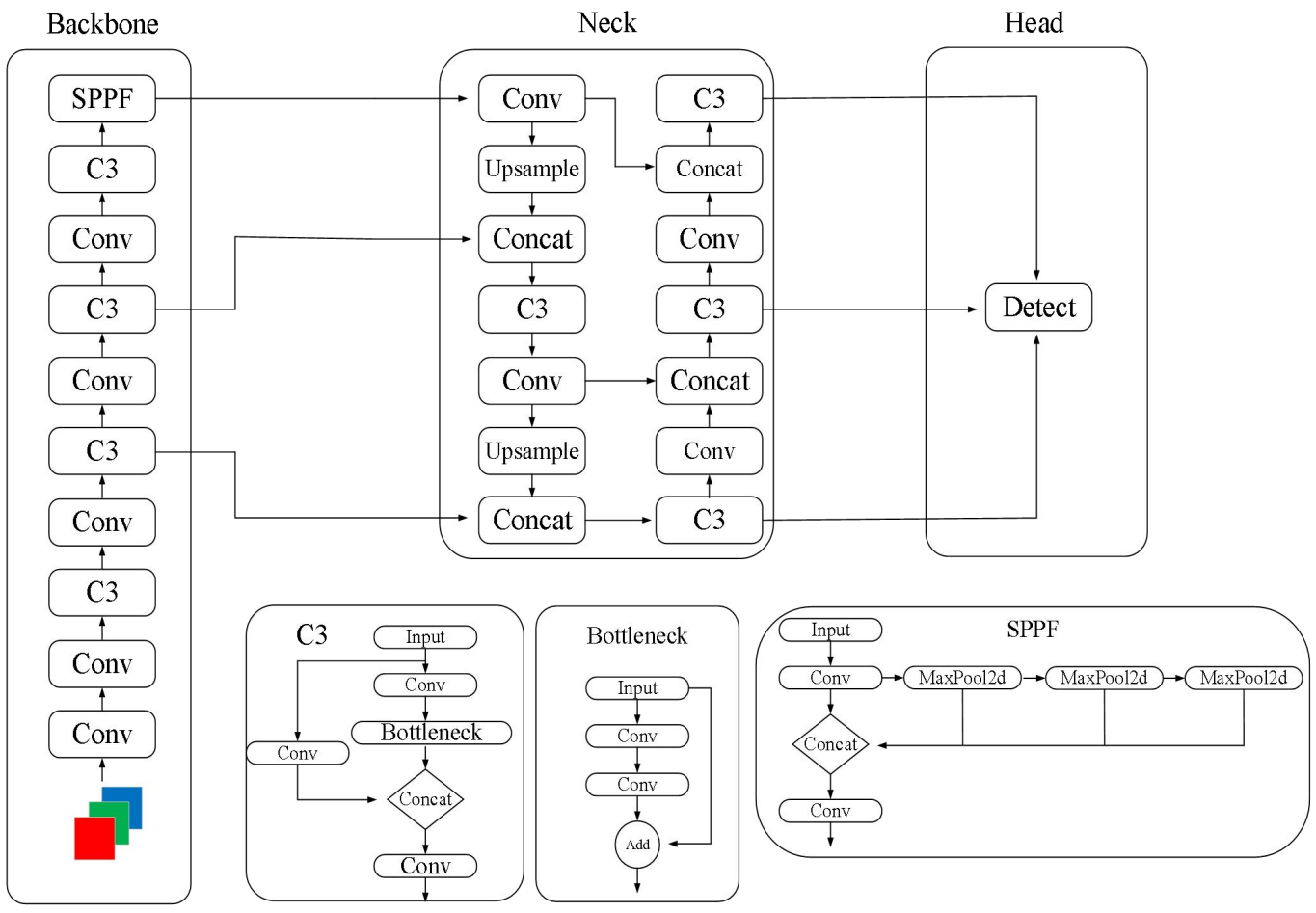

**Figure. 1.** Network Architecture of YOLOv5l.

The data input part of YOLOv5l mainly includes Mosaic data augmentation, adaptive anchor computation, and adaptive image scaling. These methods are employed to process the data and achieve a slight improvement in accuracy. Mosaic data augmentation randomly selects four images and applies operations such as flipping, cropping, and scaling to create a single image by combining them in certain proportions, This enriches the dataset with diversity, increases the number of small objects, and enhances the model's robustness. The adaptive anchor computation no longer relies on predefined anchors. Instead, before each training iteration, it calculates the most suitable anchor sizes based on the specific dataset, this is achieved by utilizing the K-means clustering algorithm to determine anchors and applying a genetic algorithm to randomly adjust their widths and heights. Adaptive image scaling is applied to the inference process of the network, while the training process still uses ordinary image scaling. The difference lies in the fact that regular scaling has more black edges and redundant information, but adaptive scaling can minimize the black edges when filling them, thereby enhancing the speed of inference.

The backbone network of YOLOv5l consists of Conv modules, C3 modules, and SPPF modules. The Conv module is composed of convolutional layers, BN (Batch Normalization) layers and activation layers, where the ReLU activation function is replaced with SiLU activation function. The C3 module is primarily inspired by the CSPNet structure, which can enhance the network's learning ability while reducing memory consumption. The SPPF module performs feature fusion across different scales to enlarge the receptive field. Unlike traditional SPP[32], the SPPF achieves the same effect by stacking smaller convolutional kernels instead of using larger ones, which helps to reduce computational complexity.

The neck of YOLOv5l consists of FPN[33] and PANet[34]. FPN performs top-down upsampling to propagate strong semantic features from deep layers to shallow layers for feature fusion. On the other hand, PANet conveys strong localization features bottom-up to provide positional information to deep feature maps, enhancing the feature fusion capability.

The head of YOLOv5l is the Detect module, which adjusts the channel dimension of the output feature maps from the Neck using convolutions. The channel size is adjusted to 3×(5+class), where 5 represents the predicted bounding box coordinates x, y, h, w, and confidence, while class represents the number of detection classes. Each grid cell generates 3 prediction boxes.The prediction part of YOLOv5l includes non-maximum suppression (NMS) processing and loss function computation. The loss function is composed of three parts: bounding box localization loss, confidence loss and classification loss. The confidence and classification losses are computed using binary cross-entropy loss, while the bounding box localization loss is computed using the CIOU Loss function.

### *3.2. Method Design*

The YOLOv5l network replaces the Focus module with a 6x6 convolutional layer, removes several concatenation operations to reduce memory overhead, but still needs improvement in feature extraction. This paper proposes a Multi-Head Attention Self-Convolution (M-sconv) module to replace the 6x6 convolutional layer, which not only improves detection accuracy but also speeds up model convergence. To address the challenge of detecting small and dense objects, this paper replaces the first C3 module in the shallow network with the Swin Transformer Block, This type of self attention is modeled based on global information, with linearly increasing computational complexity, it also performs feature fusion through sliding windows, enhancing the detection performance of small objects and improving feature extraction efficiency. To preserve the integrity of feature extraction, we introduce I-CBAM attention mechanism between the Backbone and Neck, This mechanism is improved based on the CBAM attention mechanism, and focuses on more important features to improve model accuracy. Finally, enhanced feature fusion is employed (between the 10th and 25th layers, and between the 13th and 22nd layers) for more effective communication of feature and semantic. The improved YOLOv5l model architecture is illustrated in **Figure 2**.

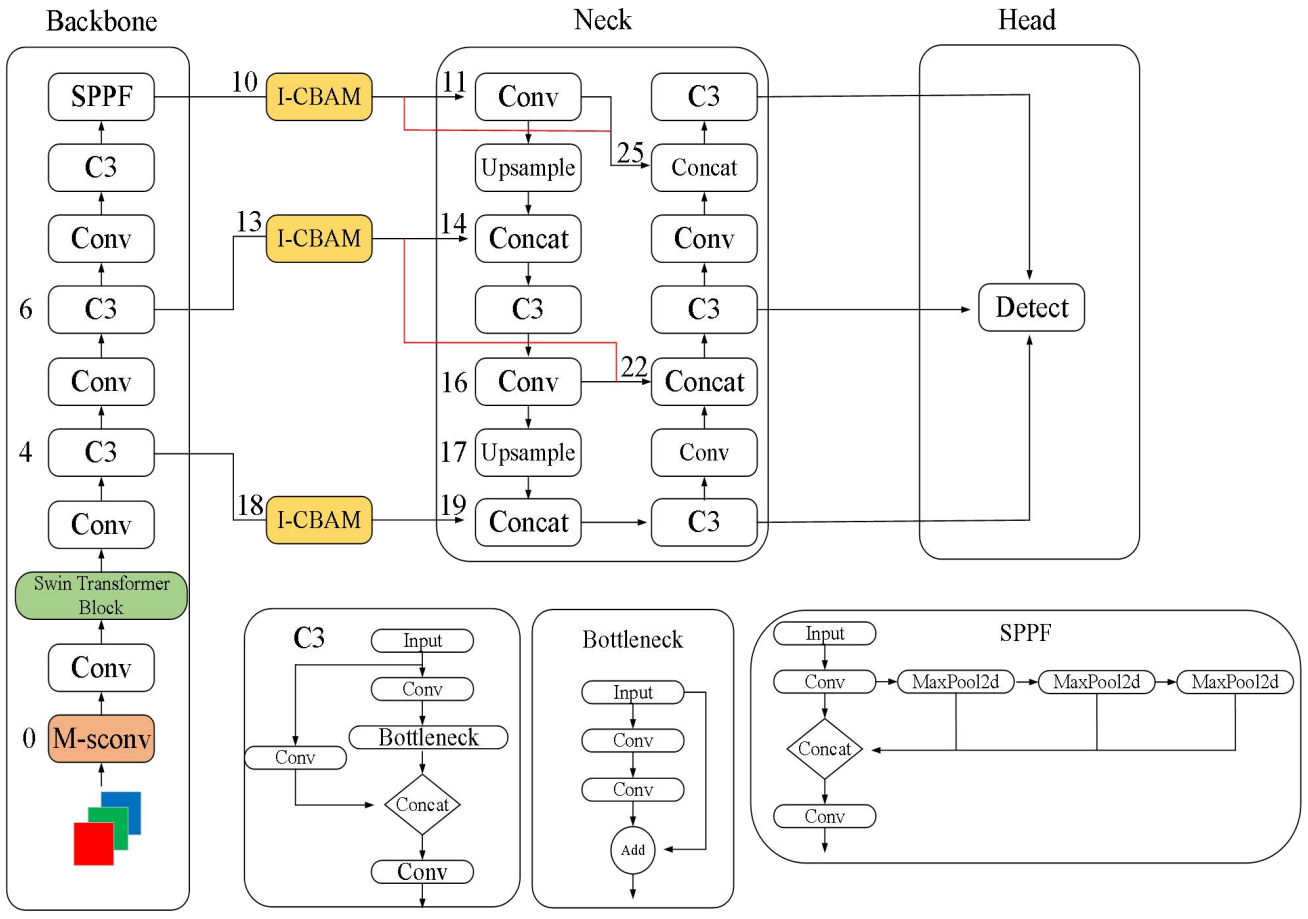

**Figure.2.** Improved YOLOv5l Model Architecture.

### *3.3. Multi-Head Attention Self-Convolution(M-sconv)*

The training weights of a regular convolutional kernel are obtained through the loss function, training with weights starting from 0 increases the complexity of the model. The proposed Multi-Head Attention Self-Convolution (M-sconv) in this paper is a module that combines CNN and Transformer[35]. It allows the integration of self-information from the input feature map into the weights of the convolutional kernel. This is achieved by utilizing global average pooling and global max pooling to capture global and salient information, which is then fused together. This approach accelerates the convergence of the entire model. Furthermore, to effectively extract features and enhance detection capability, the paper introduces the Transformer Encoder module. The model diagram of the Multi-Head Attention Self-Convolution is illustrated in **Figure 3**.

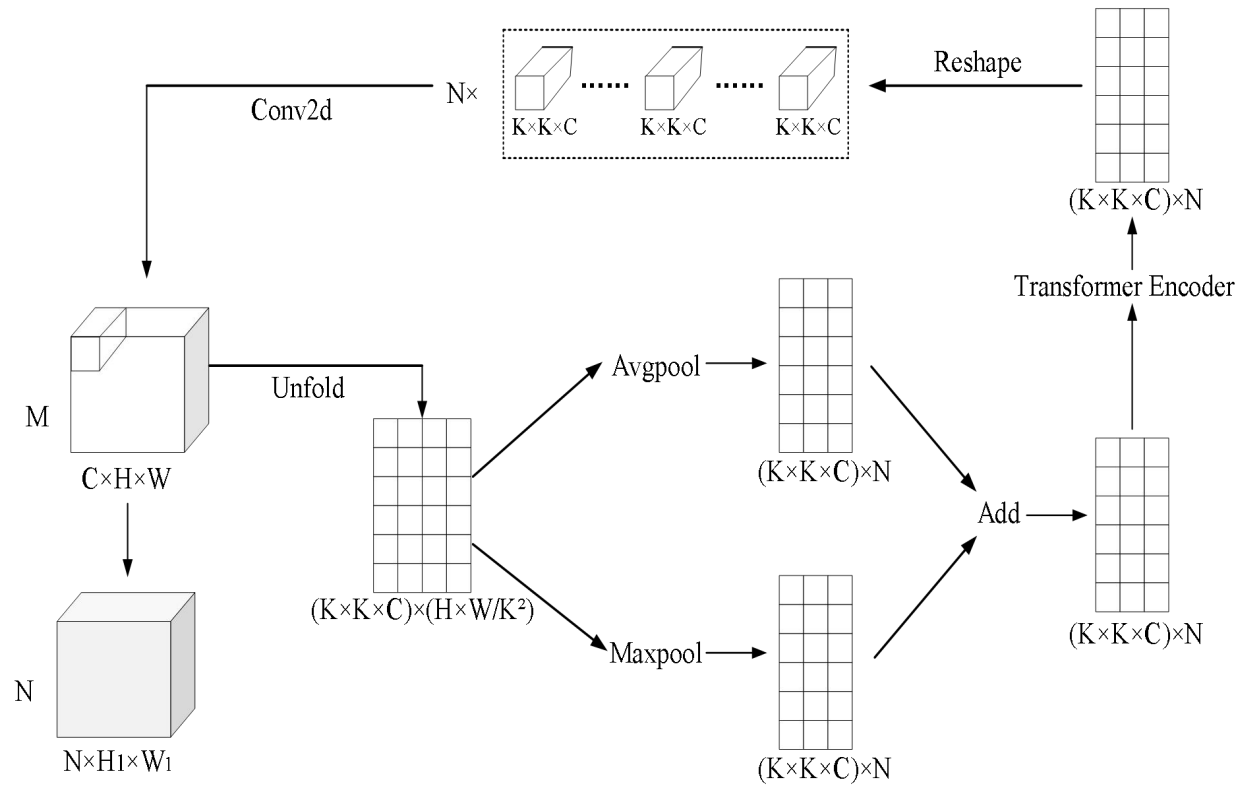

**Figure.3.** Multi-Head Attention Self-Convolution

The Multi-Head Attention Self-Convolution first unfolds by sliding windows. Each unfolding operation expands K·K pixels along the dimension of channel C. The unfolded 2D feature map has a height of K·K·C. This unfolding process is performed ($H \cdot W/K^2$) times, resulting in a width of $H \cdot W/K^2$ for the 2D feature map. Each column of the 2D feature map represents the information obtained by a convolutional kernel from the original feature map. K·K represents the size of the convolutional kernel, C corresponds to the number of channels in the kernel, and $H \cdot W/K^2$ represents the number of convolutional kernels.

Assuming the number of output channels is N, we divide the width($H \cdot W/K^2$) into N equal groups. Each group undergoes average pooling and max pooling to extract the overall information and highlight features from the feature map, and to obtain the maximum feature map and average feature map. The feature maps are then merged together, resulting in a fused feature map with a size of (K·K·C) x N.

To improve the detection capability of the model, the fused feature map is passed through a Transformer Encoder module. This module utilizes global-based self-attention and performs multi-head operations on grouped feature maps, which enhances the model's detection performance to some extent. After passing through the Transformer Encoder module, the feature map size becomes (K·K·C) x N. It is then reshaped into N convolutional kernels of size K·K with C channels. In this step, the parameters in the convolutional kernels are no longer rrandom values but obtained by integrating own information from the input feature map. Finally, these generated convolutional kernels are used to convolve the original feature map. Experimental results demonstrate that the Multi-Head Attention Self-Convolution not only improves the accuracy of network predictions, but also accelerates the model convergence.

Figure 4 presents the network architecture diagram of the Transformer Encoder. Firstly the input enters the Embedding layer to transform the words into vector representations. The resulting vectors then undergo Positional Encoding to incorporate positional information. The output of the Positional Encoding is further processed through the Multi-Head Attention layer for matrix computations. The computational steps can be described as follows:

$$head_i = Attention\left(QW_i^Q,\ KW_i^K,\ VW_i^V\right) \tag{1}$$

$$Attention\left(Q,\ K,\ V\right) = Softmax\left(QK^T/\sqrt{dk}\right)V \tag{2}$$

$$Multihead\left(Q,\ K,\ V\right) = Concat\left(head_1 \ldots\ldots,\ head_h\right)W^0 \tag{3}$$

$W_i^Q$, $W_i^K$, $W_i^V$ represent the parameter matrices of the i-th attention head, $dk$ represents the dimension of the vectors, $W^0$ is a trainable parameter matrix, in the Transformer model, the default dimension is 512. First, the dimensions are grouped using Equation 1, then single-head attention calculation is performed using Equation 2, and finally, multiple heads are concatenated together using Equation 3 to output the final result.

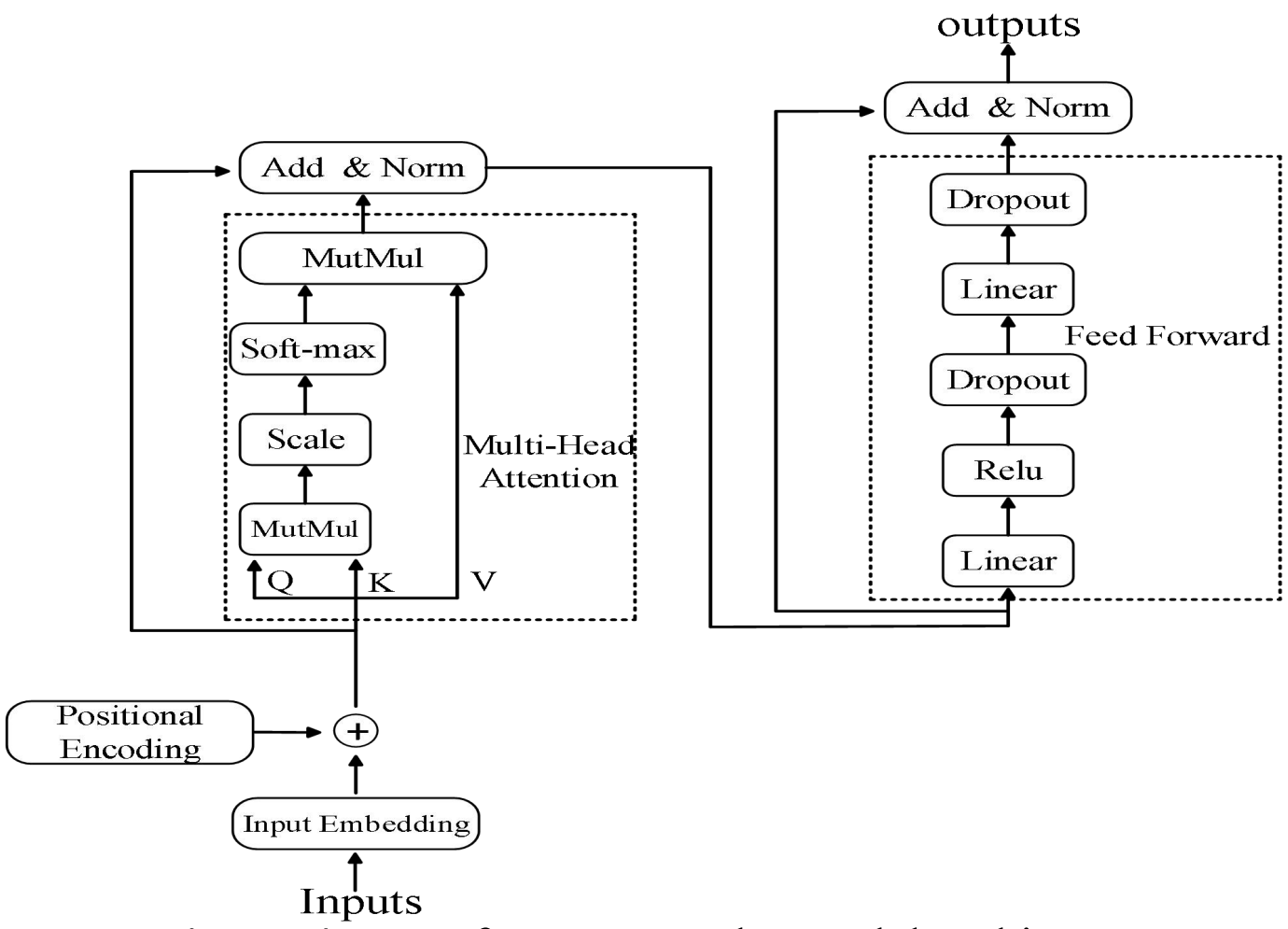

**Figure.4.** Transformer Encoder model architecture

the outputs of the Multi-Head Attention are fused through a residual structure and then normalized using LayerNorm. LayerNorm is used instead of BatchNorm for the following reason: BatchNorm normalizes Same feature(calculating mean and variance) corresponding to all samples in a batch for a given sequence. However, the lengths of sequences are different in most cases, so it is not possible to calculate mean and variance for the same feature. On the other hand, LayerNorm calculates the mean and variance across all features for the same sample, not depending on the length of the sequence. Figure 5 illustrates the difference between these two normalizations, where " Batch " represents samples in the same batch, " Seq " represents sequence length, and "Feature" represents the dimension of the vector (512). The processed output then goes through a feed-forward network and finally undergoes the residual structure and LayerNorm before being outputted.

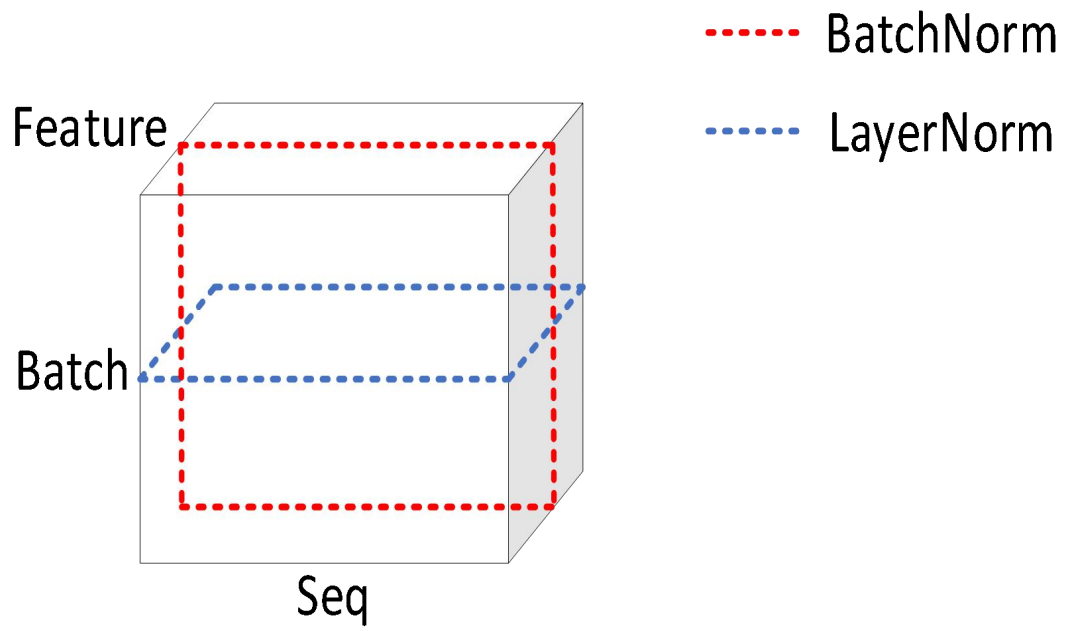

**Figure.5.** The difference between two normalization methods

In order to provide a more detailed presentation of the M-sconv module, we have described in Algorithm 3-1 how to combine the self-feature map to generate self-convolution kernels.

### *3.4. Swin Transformer Block*

The YOLOv5 model demonstrates strong generalizability, but it shows relatively weaker performance in detecting small or densely packed objects. This article chooses to replace the C3 module in the second layer of the backbone with Swin Transformer Block. Swin Transformer Block has the ability to capture global information and facilitate information propagation through sliding windows. This enables the extraction of finer details, thereby improving the detection accuracy of the model. The structure of Swin Transformer Block is illustrated in Figure 6.

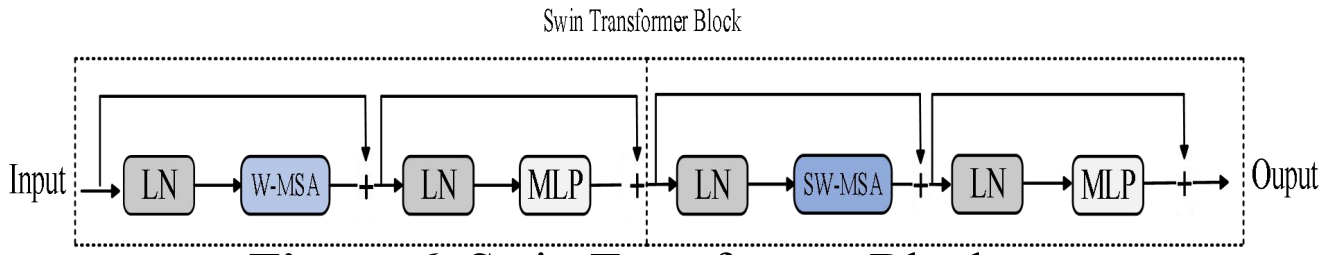

**Figure.6.** Swin Transformer Block

The Swin Transformer Block consists of two tightly coupled parts. The first part is the window-based self-attention mechanism. The input first goes through a Layer Normalization (LN) for normalization and then enters the W-MSA (Windows Multi-head Self-Attention) module. Compared to MSA, W-MSA does not send the entire image into the self attention module for calculation, instead, it introduces a sliding window mechanism that divides the input image into multiple equally sized windows. In this article, since the image size entering the W-MSA module is 160*160, the window size is set to 8x8. Then, 400 windows (20x20) are

Algorithm 2-1: Multi-Head Attention Self-Convolution Module (M-sconv)

**Algorithm 2-1：M-Sconv algorithm**

Input: Feature maps $M \in R^{C\times H\times W}$

Output: Feature maps $N \in R^{N\times H_1\times W_1}$

Initialization: Define Transformer and set parameters: input dimension N, output dimension N, num_head as 4.

1. $M \leftarrow$ Unfold $M$ // To unfold using the unfold function
2. $M_{max} \leftarrow$ Adaptive_max_poold($M$) // Adaptive Global Max Pooling
3. $M_{avg} \leftarrow$ Adaptive_avg_poold($M_{max}$) // Adaptive Global Average Pooling
4. $M_i \leftarrow$ Add($M_{max}$，$M_{avg}$) // Feature fusion of two pooled feature maps
5. $M_t \leftarrow$ Transformer($M_i$) // Using Self Attention for Feature Extraction
6. $M_r \leftarrow$ Reshape($M_t$) // Adjusting dimensions to generate self convolutional kernels
7. $N \leftarrow$Conv2d($M$，$weight\ =\ M_r$)// Using self convolutional kernels to check the original feature map for convolution

Output：$N$

calculated to determine the correlations between the pixels.

The calculation method is similar to the Transformer Encoder, with the difference being the use of relative position encoding B among the heads, as shown in formula 4.

$$Attention(Q，K，V) = Softmax\left(QK^T/\sqrt{dk} + B\right)V \qquad (4)$$

After passing through the residual connection and entering the LN layer, the input is adjusted for channel numbers through an MLP. It is then followed by another residual connection to obtain the output of the first part. The second part is the self attention of sliding windows, which has a similar structure to the first part. However, instead of the W-MSA module, it utilizes the SW-MSA (Shifted Windows Multi-Head Self-Attention). The W-MSA module computes self-attention within each window but lacks interaction between windows, making it unable to compute the similarity between pixels in different windows. Therefore, the SW-MSA module is proposed, which achieves information exchange between windows through window shifting, as shown in Figure 7(a). Due to the different patches of the moved windows, the number of windows is also more than before, increasing computational complexity. To maintain consistency with the number of windows and the number of patches per window before shifting, Cyclic Shift is applied. As shown in Figure 7(b), the first step is to move the top $\lfloor \frac{M}{2} \rfloor$ rows to the bottom and then move the leftmost $\lfloor \frac{M}{2} \rfloor$ columns to the rightmost, where M represents the window size. After window partitioning, since patches within the same window come from different regions, self-attention computation needs to be done through Masking. The Masking concept involves assigning smaller weights to the pixels from different regions within each window, ensuring that the weights become 0 after applying the softmax function, while preserving the weight values between pixels from the same region. Lastly, the Reverse Cyclic Shift is applied to restore the window positions, maintaining the relative positions and overall semantic information of the original image. Experimental results demonstrate that replacing the first C3 module with the Swin Transformer Block improves the detection accuracy.

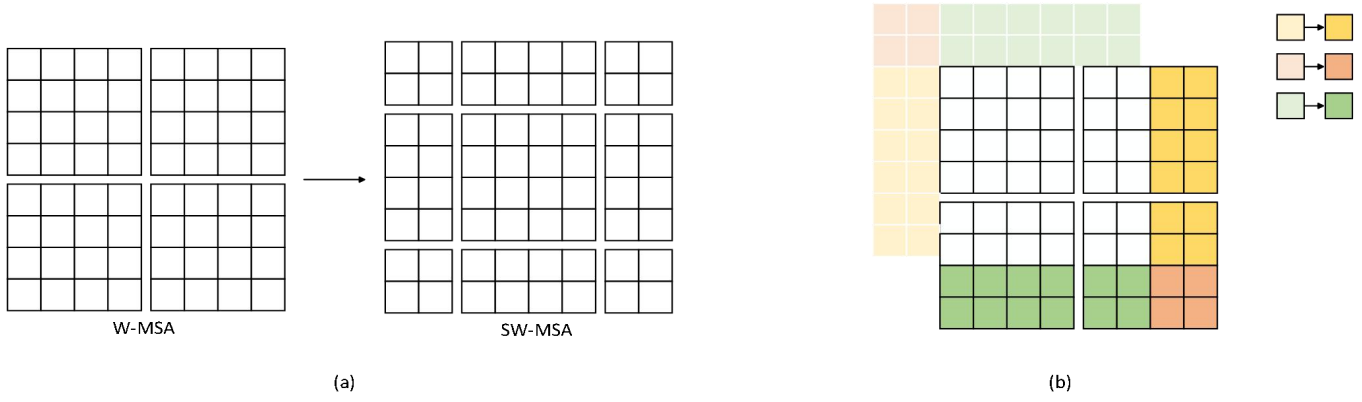

**Figure.7.** Illustration of Window Shifting

### 3.5. *Improved Attention Mechanism(I-CBAM)*

The I-CBAM attention mechanism (Improved Convolutional Block Attention Module) is an enhancement based on the CBAM attention mechanism. It improves the model's detection accuracy by enhancing focus on relevant information and reducing the importance of irrelevant information. The structure of I-CBAM is shown in Figure 8. It keeps the original structure in channel attention and The results of global average pooling and global maximum pooling are respectively processed through 7x7 convolution operations in spatial attention, increasing the receptive field while extracting features. Experimental results have shown that introducing the I-CBAM attention mechanism effectively improves the feature extraction capability. Table 3 demonstrates that the improved attention mechanism outperforms other attention mechanisms in terms of model detection accuracy.

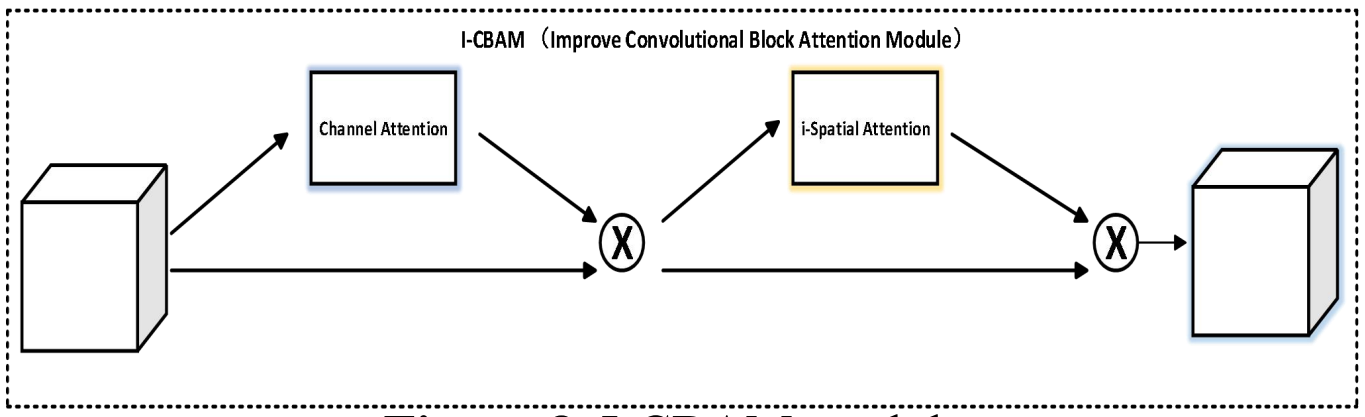

**Figure.8.** I-CBAM module

#### 3.5.1. *The channel attention module in I-CBAM(CAM)*

Channel Attention Mechanism (CAM) is inspired by the SE[36] module. As shown in Figure 9, SE compresses spatial information into channels through global average pooling. Although it has a large receptive field, it causes information loss. Therefore, the CAM module performs global average pooling and global max pooling on the input feature map separately, resulting in two 1×1×C feature vectors. These vectors then go through two shared fully connected layers, and perform feature fusion on the output information obtained from the shared fully connected layers. Finally, the Sigmoid activation function is applied to obtain the final channel attention vector. The algorithm can be expressed as follows, according to Equation 5:

$$M_1(F) = Sim\left(W_1\left(W_0(Favg^C)\right) + W_1\left(W_0(Fmax^C)\right)\right) \quad (5)$$

$Fmax^C$, $Favg^C$ represent global max pooling and global average pooling, respectively. $W_0$, $W_1$ represent the shared fully connected layers. Sigmoid represents the sigmoid activation function. $F$ represents the input feature map. $M_1$ represents the final channel attention vector.

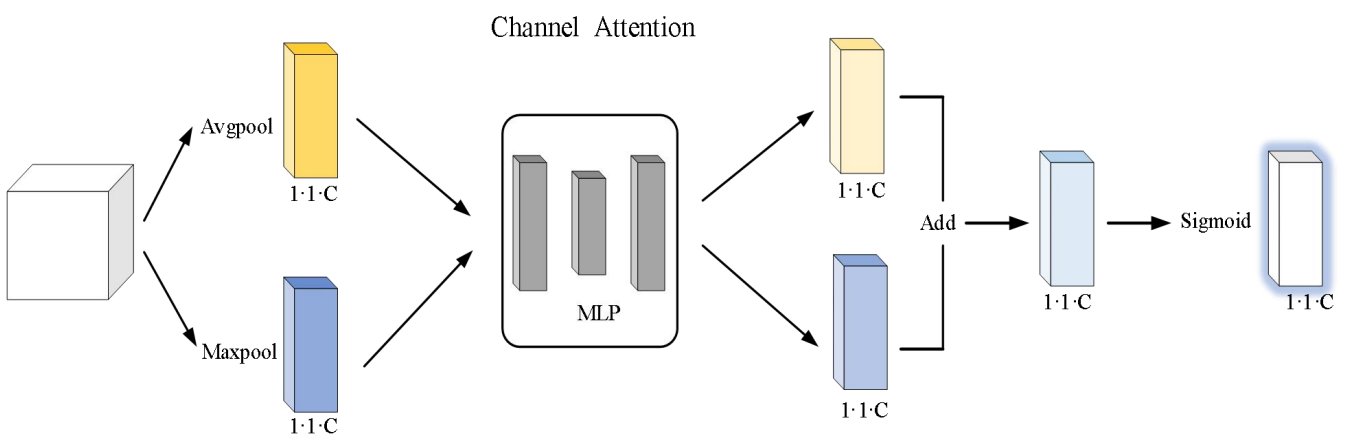

**Figure.9.** The Channel Attention Module(CAM)

### 3.5.2. *The spatial attention module in I-CBAM(I-SAM)*

The spatial attention module (I-SAM) used in this paper is inspired by the SAM module. As shown in Figure 10, the input feature map is globally average pooled and globally max pooled along the channel axis, resulting in two feature vectors of size H×W×1. To reduce feature information loss and increase receptive field, the two feature vectors are separately convolved with 7×7 convolutional kernels, and the resulting output features are then fused. Finally, the Sigmoid activation function is applied to obtain the final feature vector. The detailed description is as follows:

$$M_2(F) = Sim(\int 7 \times 7\ (Favg^C) + \int 7 \times 7\ (Fmax^C)) \quad (6)$$

$Favg^C$, $Fmax^C$ represent global average pooling and global max pooling along the channel axis, respectively. $\int 7 \times 7$ indicates the convolutional layer with a kernel size of 7×7, and $M_2$ represents the final output feature map.

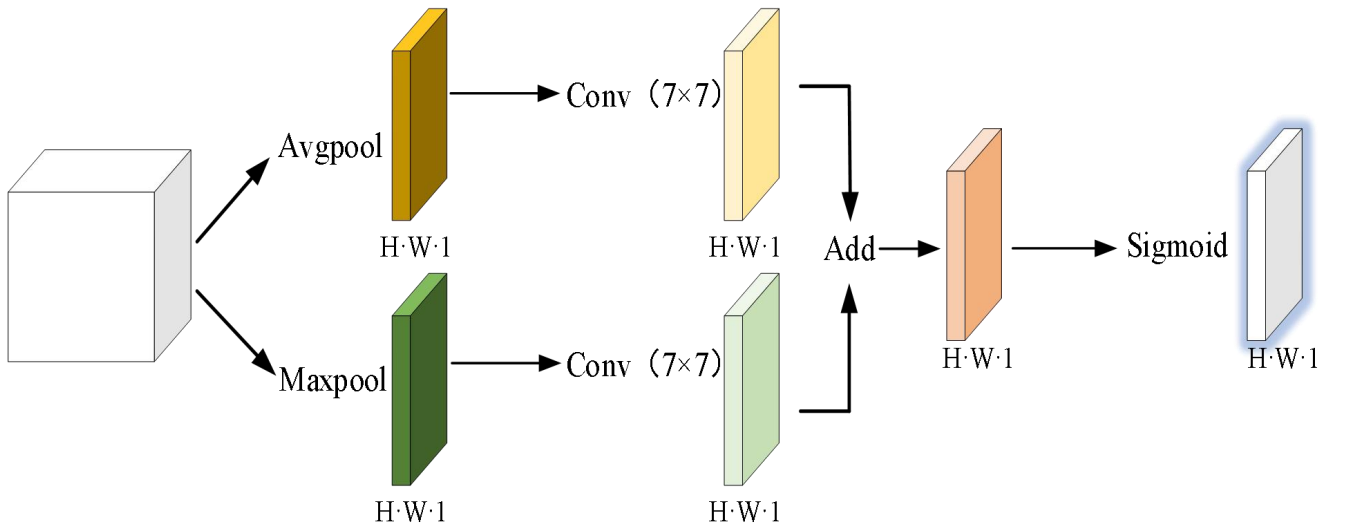

**Figure.10.** The spatial Attention Module(I-SAM)

### 3.6. *Enhanced feature fusion*

The Neck module of YOLOv5 is composed of FPN (Feature Pyramid Network) and PANet (Path Aggregation Network) structures, which effectively addresses the issue of multi-scale features at the cost of minimal computation. By fusing detail information and semantic information from different-sized features, it enhances the detection accuracy of the entire model. Based on the idea of cross-scale feature fusion, this paper enhances the feature fusion between the 10th and 25th layers, as well as the 13th and 22nd layers (The left side of the module in Figure 3 represents the number of network layers). This allows same-sized feature maps to communicate semantic and feature information more effectively, thus improving the model's accuracy. Figure 11(a) in the paper illustrates the original YOLOv5 model's cross-scale feature fusion(the number represents the number of layers in the network), where the outputs of the 13th, 18th, 11th, and 16th layers are respectively fed into the 14th, 19th, 25th, and 16th layers. In Figure 11(b), which showes the enhanced feature fusion proposed in this paper, the outputs of the 13th and 18th layers enter the 14th and 19th layers, the outputs of the 10th and 11th layers enter the 25th layers, the outputs of the 13th and 16th layers enter the 22nd layer. Experimental results indicate that using enhanced feature fusion improved detection accuracy for the model.

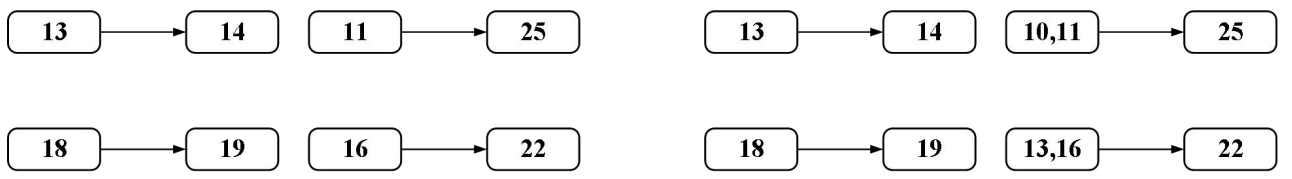

(a) Feature Fusion of the Original YOLOv5 Model (b) Enhanced feature fusion

**Figure.11.** Comparison between the improved feature fusion part and the original model

### 3.7. *Loss function*

YOLOv5l's total loss function consists of three components: class loss, object confidence loss, and bounding box loss. The class loss and object confidence loss are computed using binary cross-entropy loss, which effectively addresses the issue of slow weight update in the loss function by utilizing the Sigmoid activation function. The formula for binary cross-entropy loss is as follows:

$$C = -\frac{1}{n}\sum_x (y \ln a + (1-y)\ln(1-a)) \quad (7)$$

$y$ represents the label value, $a$ represents the predicted output value, $x$ represents the sample, $n$ represents the total number of samples.

The loss of the object bounding box usually adopts IoU Loss, GIoU Loss, DIoU Loss, CIoU Loss and other regression losses. A good localization loss needs to consider three factors: the overlap area between the predicted box and the ground truth box, the distance between their centers, and the aspect ratio. Based on these reasons, YOLOv5l chooses CIoU Loss as it is considered to be a more optimal metric. The calculation formula for CIoU Loss is represented as:

$$L_{CIoU} = 1 - CIoU \quad (8)$$

$$CIoU = IoU - (p^2(b, b^{gt})/c^2 + av) \quad (9)$$

$$v = \frac{4}{\pi^2}(arctan(w^{gt}/h^{gt}) - arctan(w/h))^2 \quad (10)$$

$$a = v/[(1 - IoU) + v] \quad (11)$$

$$IoU = A \cap B / A \cup B \quad (12)$$

$A, B$ represent the ground truth box and predicted box, $av$ represents the aspect ratio factor, $w^{gt}$, $h^{gt}$, $w$, $h$ represent the width of the ground truth box, height of the ground truth box, width of the predicted box, and height of the predicted box, $b, b^{gt}$ represent the center coordinates of the predicted box and ground truth box, $p^2(b, b^{gt})$ represents the squared distance between the center coordinates, $L_{CIoU}$ represents the bounding box loss function.

## 4. Experiment

### *4.1. Expreimental environment*

The deep learning framework used in this study is PyTorch 1.10.1, programmed in Python 3.8. The operating system is ubuntu9.4.0-1ubuntu1, The CPU model is Gen Intel(R) Core(TM) i9-13900K, and the GPU model is NVIDIA GeForce RTX 3090. The NVIDIA driver version is 525.105.17, and the CUDA version is 11.3. During the entire experimental process, the initial learning rate was set to 0.01. In the first three rounds, Warm-up learning rate was used, while cosine annealing was used to update the learning rate, The momentum factor was set to 0.937, and the weight decay factor was set to 0.0005. The batch size was set to 16, and the total number of epochs was 300.

### *4.2. Datasets*

To evaluate the performance of the network in this study, the dataset used in the experiment is sourced from the open-source Mask dataset. This dataset consists of 6,177 training images (including self-labeled small target images) and 1,782 test images. The labels in this dataset have two categories: "face" and "mask". The labels are stored in TXT files in the YOLO format, each text file containing 5 parameters, The first parameter represents the class of the ground truth box. The following 4 parameters represent the scaled center coordinates (x, y) and width (w) and height (h) of the ground truth box. Each line represents the information of a truth ground box.

### *4.3. Evaluation indicators*

In object detection, the commonly used metrics to assess the performance of a model include precision(P), Recall(R), average precision (AP), and mean average precision (mAP). P represents the ratio of predicted boxes that actually contain the object. R represents the ratio of all the true boxes that are detected. AP represents the average precision for a specific class. MAP represents the average AP across all classes. P and R are calculated using a confusion matrix, The confusion matrix is depicted in Table 1.

Table1 confusion matrix

| | prediction result | |
|---|---|---|
| actual result | positive | negative |
| positive | TP | FN |
| negative | FP | TN |

TP represents the cases where the actual class is positive and the prediction is also positive, FN represents the cases where the actual class is positive but the prediction is negative, FP represents the cases where the actual class is negative but the prediction is positive, TN represents the cases where both the actual class and the prediction are negative. The calculation formulas for P, R, AP, and mAP can be expressed as follows:

$$P = TP/(TP + FP) \tag{13}$$

$$P = TP/(TP + FN) \tag{14}$$

$$AP = \int_0^1 P(R)d(R) \tag{15}$$

$$mAP = (AP_{face} + AP_{mask})/2 \tag{16}$$

## 5. Experimental results and analysis

### *5.1. The impact of module modifications on experimental results*

The experimental detection model in this paper is based on improvements made to YOLOv5. The first and third layers of the network are replaced with the Multi-Head Attention Self-Convolution Module and the Swin Transformer Block respectively. An improved attention mechanism called I-CBAM is introduced between the Head and Neck, and enhanced feature fusion is performed on the same-sized feature maps to improve overall detection accuracy of the model. mAP(0.5) represents the average precision of two categories at an IoU value of 0.5, while mAP(0.5:0.95) represents the average precision of the two categories as the IoU value increases from 0.5 to 0.95 with a step size of 0.05. Table 2 displays the experimental results of adding each module. M-sconv represents the Multi-Head Attention Self-Convolution Module, Swin represents the Swin Transformer Block, fusion represents the improved feature fusion, and Best epoch indicates the best-performing epoch used to compare the convergence of the model after introducing the Multi-Head Attention Convolutional Module. When using the original YOLOv5l model, the mAP(0.5) and mAP(0.5:0.95) reach 91.1% and 65.5%, respectively. After adding the Multi-Head Attention Convolutional Module, mAP(0.5) increases from 91.1% to 91.3%, and mAP(0.5:0.95) improves from 65.5% to 66.3%. The Best epoch is reached 4 epochs earlier. The model not only achieves improved detection accuracy but also accelerates the convergence of the entire module. By using the Swin Transformer Block, mAP(0.5) increases from 91.1% to 91.5%, and mAP(0.5:0.95) improves from 65.5% to 66.2%. After adding the improved attention mechanism I-CBAM, mAP(0.5) increases from 91.1% to 91.6%, and mAP(0.5:0.95) improves from 65.5% to 66%. Adding the enhanced feature fusion results in both mAP(0.5) and mAP(0.5:0.95) increasing by 0.1% compared to the original model. When adding the Swin Transformer Block and the Multi-Head Attention Self-Convolution Module, mAP(0.5) increases from 91.1% to 91.8%, and mAP(0.5:0.95) improves from 65.5% to 66.2%, with the Best epoch moving up from 156 to 152. When combining three modules, including the Multi-Head Attention Self-Convolution Module, Swin Transformer Block, I-CBAM attention mechanism, mAP(0.5) increases from 91.1% to 91.9%, and mAP(0.5:0.95) improves from 65.5% to 66.3%. By simultaneously using all four modules in the YOLOv5l network, mAP(0.5) increases from 91.1% to 92.2%, an increase of 1.1%, and mAP(0.5:0.95) improves from 65.5% to 66.8%, an increase of 1.3%. The Best epoch occurs 2 epochs earlier. The experimental results demonstrate that the improved modules in this paper has improved the accuracy of mask detection while also achieving real-time object detection performance.

Table2 Add the experimental results of each module

| Network | M-sconv | Swin | I-CBAM | Fusion | P | R | mAP(0.5) | mAP(0.5:0.95) | Best epoch | Flops | FPS |
|---|---|---|---|---|---|---|---|---|---|---|---|
| YOLOv5l | | | | | 94.7% | 85.7% | 91.1% | 65.5% | 156 | 108.2 | 105 |
| | ✔ | | | | 94.9% | 86.1% | 91.3% | 66.3% | **152** | 115.3 | 98 |
| | | ✔ | | | 94.2% | 86.9% | 91.5% | 66.2% | 156 | 129.2 | 88 |
| | | | ✔ | | 94.8% | 86.6% | 91.6% | 66.0% | 158 | 108.5 | 105 |
| | | | | ✔ | 93.6% | 86.2% | 91.2% | 65.6% | 160 | 109.9 | 103 |
| | ✔ | ✔ | | | 94.9% | 86.5% | 91.8% | 66.2% | 152 | 136.2 | 83 |
| | ✔ | ✔ | ✔ | | 94.3% | 87.1% | 91.9% | 66.3% | 157 | 136.5 | 83 |
| | ✔ | ✔ | ✔ | ✔ | 94.4% | 87.7% | **92.2%** | **66.8%** | 154 | 138.1 | 82 |

Table3 Contrast the results of different attention mechanism

| Network | Model | P | R | mAP(50) | mAP(50:95) |
|---|---|---|---|---|---|
| Yolov5l | +SE | 94% | 86.7% | 91.3% | 65.8% |
| | +ECA | 93% | 87.3% | 91.3% | 65.9% |
| | +CBAM | 94.4% | 86.6% | 91.4% | 66% |
| | +I-CBAM | **94.8%** | 86.6% | **91.6%** | 66% |

Table4 Experimental comparison of different models

| Network | P | R | mAP(50) | mAP(50:95) |
|---|---|---|---|---|
| Two-stage object detection algorithms | | | | |
| Faster R-CNN | 46.0% | 63.2% | 60.2% | 33.5% |
| One-stage object detection algorithms | | | | |
| SSD+MoblilenetV2 | 95.7% | 63.6% | 77.3% | 50.3% |
| YOLOv5l | 94.7% | 85.7% | 91.1% | 65.5% |
| My-YOLOv7 | 96.2% | 85.9% | 91.7% | 65.1% |
| Ours | 94.3% | 87.1% | **92.2%** | **66.8%** |
| One-stage object detection algorithms(ancher free) | | | | |
| YOLOX-S | 94.2% | 85.8% | 90.2% | 62.3% |

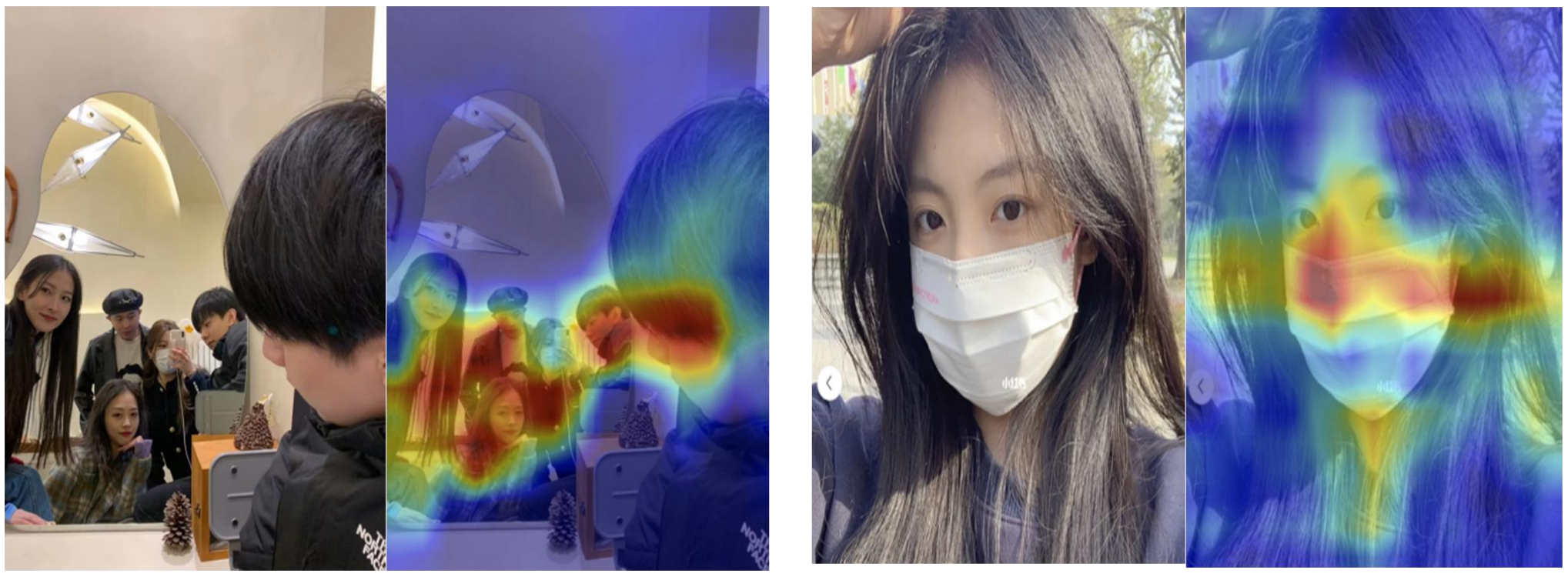

(a) (b)

**Figure.12.** visualization of heatmap

*5.2. Comparing different attention mechanisms*

To further evaluate the performance of the selected attention mechanism, this paper also conducted experiments comparing it with other attention mechanisms. Table 3 presents the impact of several different attention mechanisms on the experimental results, including SE, ECA, and CBAM. The experimental results demonstrate that the CBAM attention mechanism outperforms the others in terms of mAP(0.5) and mAP(0.5:0.95). Based on these results, this paper selects the CBAM attention mechanism for optimization. Moreover, the I-CBAM module, compared to other attention mechanisms, improves the accuracy from the original 91.4% to 91.6%.

*5.3. Heatmap*

To achieve model visualization, this paper utilizes Grad-CAM heatmaps to visually indicate the level of focus of the model on different areas of the images, enhancing the interpretability of deep learning. Figure 12(a) displays the original image of a face and the heatmap indicating the focused regions, Figure 12(b) shows the original image of a mask and the heatmap indicating the focused regions.

### *5.4. Comparison with other algorithms*

To provide a comprehensive evaluation of the performance of the model proposed in this paper, we compare it with other benchmark models. The Faster R-CNN, SSD+MobileNetV2, YOLOv5l, My-YOLOv7, and YOLOX-S models were selected for experimentation. The results are shown in Table 4. On the mask dataset, the improved algorithm in this paper achieved an accuracy of 92.2% in terms of mAP(0.5) and 66.8% in terms of mAP(0.5:0.95). These results show an improvement of over 30% compared to Faster R-CNN. Compared to algorithms in the same stage, the improved algorithm in this paper demonstrates a significant improvement in accuracy, surpassing the My-YOLOv7 version in YOLOv7. When comparing with YOLOX-S, the model used in this paper exhibits a clear advantage in accuracy.

## 6. Conclusion

In response to the issue of mask wearing in public places, this paper proposes a mask detection algorithm based on YOLOv5l. We introduce a Multi-head Attention Self-Convolution module in place of the first convolutional layer, which not only improves the ability of extracting features but also accelerates model convergence. To enhance the detection performance for dense and small objects and improve the overall accuracy of the model, we incorporate a Swin Transformer Block, based on global modeling, instead of the first C3 module. This block has the ability to capture global information and facilitates information exchange in a sliding window manner, resulting in more effective feature extraction and accuracy improvement. An improved attention I-CBAM module is introduced before the Neck to enhance attention to relevant information, achieving more effective feature extraction and improving the accuracy of the network detection. Enhanced feature fusion is performed on the feature maps of the same size to facilitate the exchange of features and semantic information, thereby improving the model's accuracy. Experimental results show that compared to the original YOLOv5 network, the optimized YOLOv5l model in this paper achieves a 1.1% and 1.3% improvement in mAP(0.5) and mAP(0.5:0.95) metrics, Future work in this area can be focused on the following aspects: 1. Reduce the model's parameters to achieve higher real-time detection performance while keeping the accuracy unchanged or slightly reduced. 2. Build a dedicated mask-wearing dataset that includes various scenarios such as front, side, and back views, to improve the model's generalization ability and robustness.

## Acknowledgements

I would like to express my heartfelt gratitude to everyone who has guided and supported me during the completion of my research paper. I am especially thankful to my teachers and my research group for their extensive knowledge, which helped me address many questions and provided me with experimental equipment. Thank you for your support and dedication. Additionally, I would also like to extend my appreciation to the scholars and experts who have contributed to the study of this topic. It is because of your foundational knowledge that I am able to move forward.